\documentclass[runningheads]{llncs}
\usepackage{graphicx}
\usepackage{lscape}
\usepackage{tabularx}
\usepackage[colorlinks=true]{hyperref}
\hypersetup{
    linkcolor=red,  
}

\begin{document}
\title{Progressive Growing of Patch Size: Resource-Efficient Curriculum Learning for Dense Prediction Tasks}
\titlerunning{Progressive Growing of Patch Size}
%
\author{Stefan M. Fischer\inst{1,2,3} \and
Lina Felsner\inst{1,2} \and
Richard Osuala\inst{1,2,4} \and
Johannes Kiechle\inst{1,2,3} \and
Daniel M. Lang\inst{1,2} \and
Jan C. Peeken\inst{1,2} \and
Julia A. Schnabel\inst{1,2,3,5}
}
\authorrunning{Fischer et al.}
%
\institute{Technical University Munich, Germany 
\and
Helmholtz Munich, Germany
\and
Munich Center of Machine Learning (MCML), Germany
\and
Universitat de Barcelona, Spain
\and
King's College London, UK
}

\maketitle              
\begin{abstract}
In this work, we introduce Progressive Growing of Patch Size, a resource-efficient implicit curriculum learning approach for dense prediction tasks. Our curriculum approach is defined by growing the patch size during model training, which gradually increases the task's difficulty. We integrated our curriculum into the nnU-Net framework and evaluated the methodology on all 10 tasks of the Medical Segmentation Decathlon. With our approach, we are able to substantially reduce runtime, computational costs, and CO$_{2}$ emissions of network training compared to classical constant patch size training. In our experiments, the curriculum approach resulted in improved convergence. We are able to outperform standard nnU-Net training, which is trained with constant patch size, in terms of Dice Score on 7 out of 10 MSD tasks while only spending roughly 50\% of the original training runtime. To the best of our knowledge, our Progressive Growing of Patch Size is the first successful employment of a sample-length curriculum in the form of patch size in the field of computer vision. Our code is publicly available at \url{https://github.com/compai-lab/2024-miccai-fischer}.

\end{abstract}
\section{Introduction}



Automatic medical image segmentation is dominated by deep learning based methods. Most research focuses on the development of new architectural concepts, introducing convolution-based \cite{ronneberger2015u,isensee2021nnu}, transformer-based \cite{dosovitskiy2020image,liu2021swin} or hybrid approaches \cite{hatamizadeh2022unetr,hatamizadeh2021swin} to improve downstream segmentation performance. In contrast, few works in the medical imaging domain focus on the actual training process of networks, and models are still trained mainly by random training data sampling. Inspired by humans, Bengio et al. \cite{bengio2009curriculum} introduced the concept of curriculum learning, teaching models to first solve simple tasks and, subsequent to that, concentrate on harder tasks. They showed that ordering samples from easy-to-hard speeds up convergence and can, therefore, also improve performance compared to training with random sampling. Those training techniques can reduce the high costs of model training in terms of time, energy, and carbon footprint \cite{selvan2022carbon}.

Different approaches have been used in the image processing field to establish such a curriculum sample ordering. Human annotations or expert knowledge can be included as problem-specific measures \cite{bengio2009curriculum,jimenez2019medical,wei2021learn}. Task difficulty can also be directly linked to sample class membership, which was utilized for fracture classification \cite{jimenez2019medical}. Another frequently applied measure in the medical domain is the inter-rater expert agreement, which considerably increases annotation costs~\cite{jimenez2019medical,wei2021learn}.

A more universal approach is rating the task difficulty automatically, or by incorporating processes that increase the task difficulty synthetically. Sample difficulty estimation can directly be computed from the network's training sample loss. Jesson et al. \cite{jesson2017cased} utilized hard-negative mining and an adaptive oversampling scheme for lung cancer segmentation. Missing modalities in MRI processing have been the motivation for Havei et al. \cite{havaei2016hemis} to train a brain tumor segmentation model that is robust against such missing data scenarios. In their curriculum learning approach, they randomly drop MRI channels with increasing probability. In the field of natural image generation, Kerras et al. \cite{karras2017progressive} defined an implicit curriculum by progressively growing their image generator and discriminator layer-wise during training. With each additional layer, the output resolution is doubled, resulting in more difficult tasks. This increased convergence speed and reduced training runtime while achieving higher model performance compared to standard training. Zhao et al. \cite{zhao2020pgu} directly transferred this idea to semantic segmentation for their task of cervical nuclei segmentation using a U-net architecture, which grows starting from the bottleneck. 

Task difficulty can also be directly defined by sample-length, which is already used in the natural language processing field in the form of sentence length~\cite{spitkovsky2010baby,zaremba2014learning,platanios2019competence}. In contrast, in the computer vision domain, the sample length, which would refer to image size or patch size, has, to the best of our knowledge, not used yet. An indirect utilization of image size is used in the Progressive Growing of GANs and its adaption to segmentation, but changes in image size are a result of the addition of network layers \cite{karras2017progressive,zhao2020pgu}. In contrast to that growing image resolution curriculum that would start segmenting a lesion from a low-resolution version of the full image, we assume that starting with segmenting a foreground object within a smaller region-of-interest patch is easier. This was our motivation to build our curriculum on image size instead of image resolution. In this work, we successfully apply the curriculum approach of growing patch size. Our main contributions in this work are:
\begin{enumerate}
    \item Introduction of Progressive Growing of Patch Size curriculum for semantic segmentation and integration into the nnU-Net framework
    \item Empirical verification of reduced computational costs and improved convergence compared to classical constant patch size training
    \item Validation of the robustness of our approach on the 10 tasks of the Medical Segmentation Decathlon (MSD) \cite{antonelli2022medical}
\end{enumerate}

\section{Methods and Materials}

\subsection{Progressive Growing of Patch Size as Curriculum}

We establish our curriculum by changing the patch size of the input volume. We assume training on smaller patches is an easier task than full images or volumes, as the frequency of foreground pixels decreases in larger patches. Furthermore, growing the patch size also results in an increased global context that can be inferred from that patch. 

\begin{figure}[t]
\centering
\includegraphics[width=\textwidth]{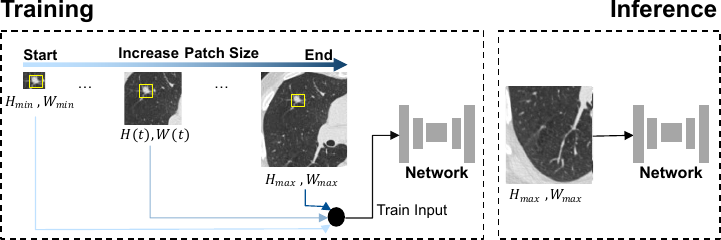}
\caption{Proposed \textbf{Progressive Growing of Patch Size} curriculum. Our curriculum is illustrated for the lung cancer segmentation with cancer bounding boxes in yellow. A fully convolutional network is able to handle inputs of different sizes. Training with our curriculum starts by training the network with minimal patch size and, with training progress, increasing the patch size until the final patch size is reached. The ratio between foreground and background voxels is bigger for small patch sizes and decreases for bigger patch sizes. In contrast, the global context that can be inferred from the patch is growing with the patch size. For inference, the maximal patch size is used.} 
\label{fig:curriculum-sketch}
\end{figure}

For our approach, we start training the network with a minimal patch size and then linearly increase the patch size to a maximal patch size during training, training at each patch size for the same number of iterations. During the model inference, the maximal possible patch size is used. A sketch of our curriculum is shown in Fig.~\ref{fig:curriculum-sketch}.

\subsection{Implementation with nnU-Net}

Fully convolutional networks are able to process inputs of varying sizes, which is the basis for our approach. While the GPU memory restricts the maximal patch size we can set, the network architecture itself restricts the minimal patch sizes that can be processed. Starting from the smallest possible patch size, we increase the patch size in the smallest possible steps to keep the similarity between different patch size stages as close as possible.


For our experiments, we use the current state-of-the-art medical segmentation network nnU-Net, which follows the U-net architecture \cite{ronneberger2015u} and is a fully self-configuring pipeline, as detailed in \cite{isensee2021nnu}. The experiments were performed on the 3D patch-based segmentation version of nnU-Net (3d\_fullres), which has shown, on average, the best performances in the MSD in \cite{isensee2019automated}. Models trained with the standard training of the nnU-Net are referred to as models trained with the \textbf{Constant Patch Size Training Scheme} (CPS). For our \textbf{Progressive Growing of Patch Size Training Scheme} (PGPS), we only change the patch size compared to CPS.

\subsection{Utilizing Smaller Patch Sizes for Bigger Batch Sizes}

Training a network with a smaller patch size reduces GPU memory consumption. Thus, PGPS has lower GPU memory consumption during most of the training compared to CPS. This enables the use of larger batch sizes in the lower patch size phases of PGPS at low GPU memory costs. With a higher batch size, we expect an increase in performance and convergence speed. Besides, we force a foreground-background ratio of one, which is the default effective ratio of nnU-Net that uses a batch size of two, a result of it's maximal patch size heuristic.

\section{Experiments and Results}

\subsection{Lung Cancer Segmentation}

\subsubsection{Experiments} We compare nnU-Net instances trained via CPS (default nnU-Net) and nnU-Net instances trained via our PGPS on the MSD task of lung cancer segmentation. To evaluate the effect of growing patch size, we also train models with \textbf{Random Patch Size Sampling} (RPSS). For that, at each training iteration, a random patch size of all patch sizes used for PGPS is picked, and a training iteration with the chosen patch size is performed. At testing time, all configurations use the same patch size, which is the nnU-Net default patch size. 

To evaluate the convergence properties of the three different curricula, we train models for different numbers of training iterations per epoch. The standard nnU-Net 250 iterations per epoch refers to 100\% scenario. Moreover, we also train models with only 10\%, 25\%, 50\% of iterations per epoch. For each model training, we track the average training runtime and compute the mean Dice Score and the general number of voxels shown to the network during training.

Performance regarding training iterations, iterated voxels, and overall training runtime for trained models are plotted in Fig.~\ref{fig:lung-cancer-seg}. Exact values are given in the Supplemental Table \textcolor{red}{3}. Performance of the 100\% iterations per epoch training of the CPS are taken from \cite{isensee2019automated}.

\begin{figure}[t!]
\includegraphics{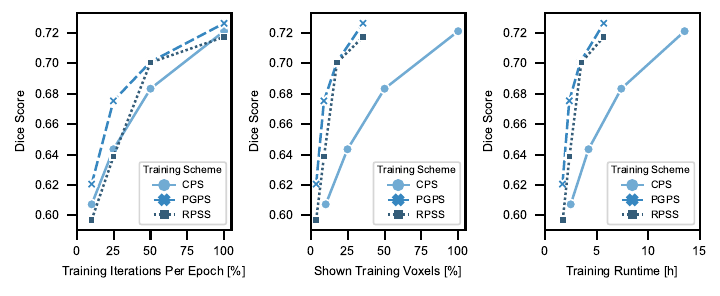}
\caption{Segmentation performance of models trained with \textbf{Progressive Growing of Patch Size} (PGPS), \textbf{Random Patch Size Sampling} (RPSS) and \textbf{Constant Patch Size} (CPS) for different numbers of training iterations per epoch on the MSD Lung Cancer Task. Models were trained with 10\%, 25\%, 50\%, and 100\% of the training iterations per epoch, while 100\% represents the default 250 iterations per epoch of nnU-Net. Dice Scores are averaged over a 5-fold cross-validation. All three plots are referring to the same models but are evaluated regarding different measures.} 
\label{fig:lung-cancer-seg}
\end{figure}

\begin{table}[t!]
\caption{Performance of \textbf{Constant Patch Size Training Scheme} (CPS), \textbf{Progressive Growing of Patch Size Training Scheme} (PGPS), and PGPS with increased batch size (PGPS+) on Medical Segmentation Decathlon. CPS refers to standard nnU-Net training. (\textbf{Dice Score:} Evaluated in 5-fold Cross-Validation as in \cite{isensee2019automated}; \textbf{P-value:} One-sided paired T-test on validation Dice Scores of all samples against CPS (* refers to significant difference [P-value $<$ 0.05]); \textbf{Runtime:} Training runtime per fold on one NVIDIA A100-SXM4-80GB GPU; \textbf{Voxels Shown:} total number of voxels iterated during training normalized to CPS training; \textbf{CO$_{2}$-eq:} mean CO$_{2}$-equivalent for training a single nnU-Net instance)}
\label{tab:MSD-performance}
\centering
\begin{scriptsize}
\begin{tabular}{|wl{2.0cm}|wl{1.1cm}||wr{1.5cm}|wr{1.5cm}|wr{1.5cm}|wr{1.8cm}|wr{1.5cm}|}
\hline
&&&&&& \\[-0.8em]
MSD Task & Scheme & Dice Score & P-value & Runtime & Voxels Shown & CO$_{2}$-eq\\
\hline
\hline
 & CPS & 0.7411 & -\phantom{*} & 27.53 h & 100.00 \% & 10.44 kg \\
Brain & PGPS & 0.7412 & 0.5972\phantom{*} & 11.29 h & 38.08 \% & 4.31 kg \\
& PGPS+ & \textbf{0.7421}& 0.1173\phantom{*} & 13.83 h & 51.78 \% & 5.21 kg \\
\hline
 & CPS & \textbf{0.9328} & -\phantom{*} & 15.04 h & 100.00 \% & 6.24 kg\\
Heart & PGPS & 0.9321 & 0.2567\phantom{*} & 6.44 h & 35.35 \% & 2.64 kg\\
& PGPS+ & 0.9328 & 0.4763\phantom{*} & 7.06 h & 39.35 \% & 2.89 kg\\
\hline
 & CPS & \textbf{0.7971} & -\phantom{*} & 11.09 h & 100.00 \% & 4.70 kg\\
Liver & PGPS & 0.7891 & 0.0992\phantom{*} & 5.14 h & 38.08 \% & 2.14 kg\\
& PGPS+ & 0.7938 & 0.2369\phantom{*} & 6.70 h & 51.78 \% & 2.77 kg\\
\hline
 & CPS & 0.8891 & -\phantom{*} & 2.14 h & 100.00 \% & 0.90 kg\\
Hippocampus & PGPS & \textbf{0.8911} & 0.0073* & 1.37 h & 35.05 \% & 0.54 kg\\
& PGPS+ & 0.8907 & 0.0343* & 1.42 h & 35.23 \% & 0.56 kg\\
\hline
 & CPS & 0.7537 & -\phantom{*} & 10.14 h & 100.00 \% & 4.33 kg\\
Prostate & PGPS & \textbf{0.7566} & 0.4021\phantom{*} & 5.18 h & 30.32 \% & 2.11 kg\\
& PGPS+ & 0.7531 & 0.4679\phantom{*} & 5.75 h & 36.33 \% & 2.31 kg\\
\hline
 & CPS & 0.7211 & -\phantom{*} & 13.55 h & 100.00 \% & 5.59 kg\\
Lung & PGPS & 0.7263 & 0.3163\phantom{*} & 5.70 h & 35.35 \% & 2.35 kg\\
& PGPS+ & \textbf{0.7333} & 0.1484\phantom{*} & 6.13 h & 39.35 \% & 2.55 kg\\
\hline
 & CPS & 0.6745 & -\phantom{*} & 11.01 h & 100.00 \% & 4.68 kg\\
Pancreas & PGPS & \textbf{0.6824} & 0.0681\phantom{*} & 5.11 h & 33.93 \% & 2.07 kg\\
& PGPS+ & 0.6822 & 0.0738\phantom{*} & 5.19 h & 35.92 \% & 2.30 kg\\

\hline
 & CPS & 0.6837 & -\phantom{*} & 15.12 h & 100.00 \% & 6.06 kg\\
Hepatic Vessel & PGPS & 0.6782 & 0.0210* & 6.18 h & 36.19 \% & 2.51 kg\\
& PGPS+ & \textbf{0.6871}& 0.0633\phantom{*} & 6.65 h & 40.36 \% & 2.72 kg\\
\hline
 & CPS & 0.9638 & -\phantom{*} & 11.24 h & 100.00 \% & 4.76 kg\\
Spleen & PGPS & 0.9621 & 0.3803\phantom{*} & 5.22 h & 36.30 \% & 2.15 kg\\
& PGPS+ & \textbf{0.9654} & 0.4147\phantom{*} & 5.37 h & 45.81 \% & 2.31 kg\\
\hline
 & CPS & 0.4553 & -\phantom{*} & 10.94 h & 100.00 \% & 3.77 kg\\
Colon & PGPS & 0.4925 & 0.0046* & 4.36 h & 32.21 \% & 1.69 kg\\
& PGPS+ & \textbf{0.4967} & 0.0087* & 4.50 h & 34.55 \% & 1.77 kg\\
\hline
\end{tabular}
\end{scriptsize}

\end{table}

\subsubsection{Results} Focusing on the number of training iterations per epoch (Fig.~\ref{fig:lung-cancer-seg} [left]), PGPS models outperformed RPSS and CPS models in Dice Score for each trained configuration. RPSS is outperformed by CPS and PGPS for all training configurations except for 50\%, where it outperforms CPS. When comparing the same models regarding training efficiency, in terms of shown voxels during training (Fig.~\ref{fig:lung-cancer-seg} [center]) and training runtime (Fig.~\ref{fig:lung-cancer-seg} [right]), PGPS and RPSS have drastically increased efficiency. The number of shown training voxels for PGPS and RPSS is substantially reduced for each configuration to only 35.5\% of the original CPS training amount. Training runtime is reflecting the trend of voxels shown. Training with 100\% iterations per epoch is only running for 5.7 hours for PGPS instead of 13.55 hours for CPS.

\subsection{Medical Segmentation Decathlon}

\subsubsection{Experiments} To evaluate the robustness of PGPS on a variety of medical segmentation tasks, we tested our curriculum learning approach on all ten different tasks of the MSD. Detailed descriptions of the tasks are given in \cite{antonelli2022medical}. Various anatomies have to be segmented, including tumors, vessels, and healthy organs on single or multi-modality input, covering CT and MRI images. The MSD covers binary and multi-class segmentation tasks and dataset sample sizes ranging from 30 to 750 samples. Dice Score, runtime, and total number of voxels shown during training are computed over a 5-fold cross-validation as in \cite{isensee2019automated}. Besides, we adopt \textit{CodeCarbon} \cite{schmidt2021codecarbon} to track the CO$_{2}$-equivalents of our training process. Dice Scores of CPS models are taken from \cite{isensee2019automated}. Because of that, we extrapolated the runtime values for CPS from the runtimes of PGPS by multiplying the averaged measured time for one epoch of the maximal patch size times the number of training epochs. Used nnU-Net model architectures and training hyperparameters can be seen for each MSD task in the Appendix of \cite{isensee2019automated}. Concrete patch sizes used for PGPS are given in the Supplemental Table \textcolor{red}{3}. Furthermore, we also compute P-values for paired one-sided T-tests to assess if the performance differs significantly. For that, we pair validation Dice Scores of each sample over all cross-validation splits regarding CPS and PGPS.

Besides training with PGPS, we also repeat experiments with increased batch size. In our approach, we increase the batch size of the current patch size stage to the extent that the total number of voxels in an input tensor is equal to or lower than that of the following patch size stage. In this way, we avoid drops in the number of input voxels per iteration. We refer to this adaption of PGPS as PGPS+ in the later text. The concrete patch and batch sizes used for PGPS and PGPS+ are given in the Supplemental Table \textcolor{red}{3}.




\subsubsection{Results} 
\begin{figure}[t!]
\includegraphics[width=\textwidth]{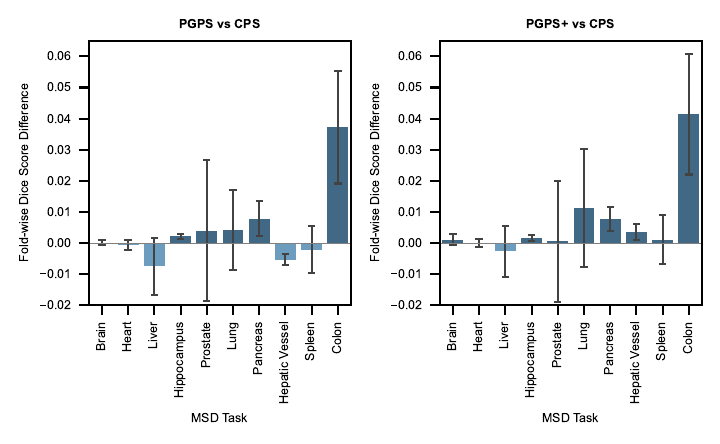}
\caption{Performance of \textbf{Progressive Growing of Patch Size} (PGPS), PGPS with increased batch size (PGPS+) and \textbf{Constant Patch Size} (CPS) on the MSD tasks. Here, we show the fold-wise mean difference in Dice Score and its standard error of the 5-fold cross-validation. A positive difference represents wins of the PGPS curricula over the CPS training (here marked in dark blue). On this measure, PGPS outperforms CPS in 6 of 10 tasks, and PGPS+ outperforms CPS in 8 of 10 tasks.} 
\label{fig:msd_performance_barplots}
\end{figure}

Results on the MSD tasks are given in Table~\ref{tab:MSD-performance}. In Fig.~\ref{fig:msd_performance_barplots}, we plotted the fold-wise Dice Score differences between our new training curricula and standard CPS. 
On averaged Dice Scores over the 5-fold cross-validation, the PGPS models outperform CPS models in 6 out of 10 tasks. In the hippocampus and the colon cancer task, PGPS outperforms CPS significantly on the sample-based T-Test (P-value $<$ 0.05), while PGPS is outperformed significantly by CPS on the hepatic vessel task. When PGPS outperformed CPS, it achieved an average fold-wise increase of $1.8 \% \pm 2.9\%$ in Dice Score compared to CPS, while for tasks CPS outperformed PGPS, there was only an average fold-wise difference of $0.5 \% \pm 0.4\%$ in performance. Full training of the nnU-Net with our PGPS curriculum reduces the runtime on average to $46.09\% \pm 6.8\%$ compared to CPS (default nnU-Net training) by only iterating $35.09\% \pm 2.30\%$ of the CPS' seen voxels and reducing CO$_{2}$ equivalents to $48.22\% \pm 9.34\%$.

With the additional adjustment of batch size for PGPS+, we are able to outperform CPS in 7 out of 10 MSD tasks on the mean of fold-wise Dice Scores. Sample-based significant performance increases were observed for the hippocampus, pancreas, and colon cancer tasks. When PGPS+ outperformed CPS, it achieved an average increase of $1.7 \% \pm 3.1\%$ in fold-wise average Dice Score compared to CPS, while when it lost, there was only an average difference of $0.2 \% \pm 0.2\%$. 
Comparing the performance differences between each fold, our PGPS+ even outperforms CPS in 8 out of 10 tasks, which can be seen in Fig.~\ref{fig:msd_performance_barplots}. The model training only takes up an average of $50.59\% \pm 7.6\%$ of the original time of CPS, while iterated voxels are reduced to $41.05\% \pm 6.20\%$ and CO$_{2}$ equivalents to $50.59\% \pm 5.55\%$.

\section{Discussion and Conclusion}

We introduced a novel curriculum learning approach based on increasing the patch size during training and compared it with constant patch size training. 

Our curriculum approach showed improved convergence in the experiments, as it significantly outperforms classical CPS on the MSD most difficult task of colon cancer by a large margin and furthermore shows improved or comparable performance in the other tasks. Better convergence is a proven property of optimal curricula \cite{weinshall2018curriculum}. In the lung cancer experiment, we showed that the performance gain per training voxel is drastically increased compared to CPS. This was observed for all different training configurations, differing in the length of network training. The beneficial impact of our curriculum is based on the ordering of patches by growing size, which we assume results in an efficient, easy-to-hard order. Furthermore, PGPS also outperforms random ordering (RPSS), validating the impact of the patch size ordering. We hypothesize that with growing patch size, we ask the network subsequently more difficult questions. In contrast, Li et al. \cite{li2020new} used only two different patch sizes during training as a comparison method to their proposed curriculum. Their results showed a performance drop of their \textit{patch-to-whole} training over CPS training. We assume that our approach benefits from maximal patch size stages.


Moreover, our curriculum can utilize larger batch sizes within the same GPU hardware restrictions. In our experiments, this resulted in better or comparable performance for PGPS+ compared to classical CPS training, as higher batch sizes generally can lead to better convergence and performance. 

Besides the improved convergence, our proposed curriculum also drastically reduced the training runtime and CO$_{2}$ emissions to roughly half of the original values for PGPS and PGPS+. This is a result of the cheaper operations due to smaller input sizes. Minimal training runtime is defined by the total number of seen voxels during training and depends highly on the technical implementation. 

We acknowledge that this is a proof of concept work, relying on the hyperparameters of nnU-Net. More experiments are needed to explore the different effects of using another minimal patch size, number of patch size stages, batch size, normalization, and training scheduler. Furthermore, our approach is not limited to fully convolutional networks but could also be applied to transformers~\cite{dosovitskiy2020image,liu2021swin} and hybrid model architectures \cite{hatamizadeh2021swin,hatamizadeh2022unetr} as well as object detection tasks.


Overall, the proposed Progressive Growing of Patch Size is a resource-efficient curriculum strategy that drastically reduces the training runtime and furthermore can also lead to improved convergence. The performance gain per voxel is substantially higher than for classical constant patch size training, supporting our hypothesis of better convergence. Improving the efficiency of network training is particularly important considering the growing carbon footprint of training deep learning models \cite{selvan2022carbon}. 







%
%
%
\bibliographystyle{splncs04}
\bibliography{sample}

\renewcommand{\tablename}{Supplemental Table}

%

\begin{table}[t]
\caption{Performance of \textbf{Constant Patch Size Training Scheme} (CPS) and \textbf{Progressive Growing of Patch Size Training Scheme} (PGPS) and \textbf{Random Patch Size Sampling Training Scheme} (RPSS) on Task06 Lung Cancer of Medical Segmentation Decathlon for different numbers of training iterations. (\textbf{Iterations Per Epoch}: Number of training iterations per epoch normalized to default training iterations of nnU-Net; \textbf{Dice Score}: Evaluated in 5-fold Cross-Validation; \textbf{Runtime}: Training runtime per fold on one NVIDIA A100-SXM4-80GB GPU; \textbf{Voxels Shown}: Total number of voxels iterated during training normalized to CPS training)}
\label{tab:lung-cancer-seg}
\centering
\begin{scriptsize}
\begin{tabular}{|l|l||r|r|r|}

\hline
Iterations Per Epoch & Training Scheme & Dice Score  & Runtime  & Voxels Shown  \\
\hline
\hline
10\% & CPS & 0.6073 & 2.49 h& 10.0 \% \\
& RPSS & 0.5971 & 1.72 h& $\sim$ 3.5 \% \\
& PGPS & \textbf{0.6608} & 1.69 h& 3.5 \% \\
\hline

25\% & CPS & 0.6435 & 4.22 h& 25.0 \% \\
& RPSS & 0.6386 & 2.37 h& $\sim$ 8.8 \% \\
& PGPS & \textbf{0.6753} & 2.35 h& 8.8 \% \\
\hline

50\% & CPS & 0.6833 & 7.40 h& 50.0 \% \\
& RPSS & 0.7005 & 3.52 h& $\sim$ 17.7 \% \\
& PGPS & \textbf{0.7009} & 3.43 h& 17.7 \% \\
\hline

100\% & CPS & 0.7211 & 13.55 h& 100.0 \% \\
& RPSS & 0.7173 & 5.69 h& $\sim$ 35.4 \% \\
& PGPS & \textbf{0.7263} & 5.70 h& 35.4 \% \\
\hline
\end{tabular}
\end{scriptsize}
\end{table}

\begin{table}
\caption{All training input tensor sizes for nnU-Net model training with Progressive Growing of Patch Size Curriculum on the Medical Segmentation Decathlon tasks. The input is given as batch size * width * height * depth. These are the values used for PGPS+. The raw PGPS uses a constant batch size for all patch size stages, which is here given for the final patch size. The maximum patch size during model training is also used for inference.}
\label{tab:PGPS-patch-sizes}
\begin{scriptsize}
\begin{tabular}{|p{1.3cm}||p{2cm}|p{2cm}|p{2cm}|p{2.1cm}|p{2cm}|}
\hline
 & Brain & Heart & Liver & Hippocampus & Prostate \\
\hline
\hline
1. Stage & 24*64*32*32 & 24*32*32*32 & 24*64*32*32 & 24*16*8*8 & 24*8*64*64 \\

2. Stage & 12*64*64*32 & 12*32*64*32 & 12*64*64*32 & 12*16*16*8 & 12*8*128*64 \\

3. Stage & 6*64*64*64 & 6*32*64*64 & 6*64*64*64 & 9*16*16*16 & 6*8*128*128\\

4. Stage & 4*96*64*64 & 4*48*64*64 & 4*96*64*64 & 9*24*16*16 & 4*12*128*128\\

5. Stage & 3*96*96*64 & 3*48*96*64 & 3*96*96*64 & 9*24*24*16 & 3*12*192*128\\

6. Stage & 2*96*96*96 & 2*48*96*96 & 2*96*96*96 & 9*24*24*24 & 2*12*192*192\\

7. Stage & 2*128*96*96 & 2*64*96*96 & 2*128*96*96 & 9*32*24*24 & 2*16*192*192\\

8. Stage & 2*128*128*96 & 2*64*128*96 & 2*128*128*96 & 9*32*32*24 & 2*16*256*192\\

9. Stage & 2*128*128*128 & 2*64*128*128 & 2*128*128*128 & 9*32*32*32 & 2*16*256*256 \\

10. Stage & - & 2*80*128*128 & - & 9*40*32*32 & 2*20*256*256\\

11. Stage & - & 2*80*160*128 & - & 9*40*40*32 & 2*20*320*256\\

12. Stage & - & 2*80*160*160 & - & 9*40*40*40 & - \\

13. Stage & - & 2*80*192*160 & - & 9*40*48*40 & -\\

14. Stage & - & - & - & 9*40*56*40 & -\\

15. Stage & - & - & - & - & - \\
\hline
\end{tabular}
\end{scriptsize}


\centering
\begin{scriptsize}
\begin{tabular}{|p{1.3cm}||p{2cm}|p{2cm}|p{2cm}|p{2.1cm}|p{2cm}|}
\hline
 & Lung & Pancreas & Hepatic Vessel & Spleen & Colon \\
\hline
\hline
1. Stage & 24*32*32*32 & 24*16*32*32 & 24*32*32*32 & 24*32*32*32 & 24*16*32*32\\

2. Stage & 12*32*64*32 & 12*16*64*32 & 12*32*64*32 & 12*32*64*32 & 12*16*64*32\\

3. Stage & 6*32*64*64 & 6*16*64*64 & 6*32*64*64 & 6*32*64*64 & 6*16*64*64\\

4. Stage & 4*48*64*64 & 4*24*64*64 & 4*48*64*64 & 4*48*64*64 & 4*24*64*64\\

5. Stage & 3*48*96*64 & 3*24*96*64 & 3*48*96*64 & 3*48*96*64 & 3*24*96*64\\

6. Stage & 2*48*96*96 & 2*24*96*96 & 2*48*96*96 & 2*48*96*96 & 2*24*96*96\\

7. Stage & 2*64*96*96 & 2*32*96*96 & 2*64*96*96 & 2*64*96*96 & 2*32*96*96\\

8. Stage & 2*64*128*96 & 2*32*128*96 & 2*64*128*96 & 2*64*128*96 & 2*32*128*96\\

9. Stage & 2*64*128*128 & 2*32*128*128 & 2*64*128*128 & 2*64*128*128 & 2*32*128*128\\

10. Stage & 2*80*128*128 & 2*40*128*128 & 2*64*160*128 & 2*64*160*128 & 2*40*128*128\\

11. Stage & 2*80*160*128 & 2*40*160*128 & 2*64*160*160 & 2*64*160*160 & 2*40*160*128\\

12. Stage & 2*80*160*160 & 2*40*160*160 & 2*64*192*160 & 2*64*192*160 & 2*40*160*160\\

13. Stage & 2*80*192*160 & 2*40*192*160 & 2*64*192*192 & - & 2*48*160*160\\

14. Stage & - & 2*40*192*192 & - & - & 2*48*192*160\\

15. Stage & - & 2*40*224*192 & - & - & 2*56*192*160 \\
\hline
\end{tabular}
\end{scriptsize}

\end{table}

\end{document}